# StackGAN: Facial image generation optimizations

Belhiti, B., Milushev, J., Gupta, A., Breedis, J., Dinh, J., Pisel, J.R.[2], Pyrcz, M.J.[3,4]

[1]College of Natural Sciences, The University of Texas at Austin, Austin, Texas

[2]Paul M. Rady School of Computer Science and Engineering, The University of Colorado at Boulder, Gunnison, Colorado

[3]Cockrell School of Engineering, The University of Texas at Austin, Austin, Texas

[4]Jackson School of Geosciences, The University of Texas at Austin, Austin, Texas

## Executive Summary

Generative adversarial networks (GANs), used to generate synthetic data, often fail to adequately reproduce higher resolution image patterns. In our novel StackGAN framework, generation of facial images is split among multiple stages, which each generate a different feature. This complexity reduction can be used for higher quality generation to construct more realistic final images. Our approach achieves subpar results when compared to industry standards using the state-of-the-art GAN evaluation metric. However, our proposed method has the potential to improve and generalize well by incorporating measures to reduce overfitting and noise. Since our design is not domain specific, our model can be used across a variety of datasets and problems, resulting in more effective generation of synthetic high resolution images in a multitude of fields. Ultimately, we learned the importance of integrating diverse training-images and that small changes in GAN architecture, such as the inclusion of an additional stage, can result in large model performance swings.

## Abstract

Current state-of-the-art photorealistic generators are computationally expensive, involve unstable training processes, and have real and synthetic distributions that are dissimilar in higher-dimensional spaces. To solve these issues, we propose a variant of the StackGAN architecture. The new architecture incorporates conditional generators to construct an image in many stages. In our model, we generate grayscale facial images in two different stages: noise to edges (stage one) and edges to grayscale (stage two). Our model is trained with the CelebA facial image dataset and achieved a Fréchet Inception Distance (FID) score of 73 for edge images and a score of 59 for grayscale images generated using the synthetic edge images. Although our model achieved subpar results in relation to state-of-the-art models, dropout layers could reduce the overfitting in our conditional mapping. Additionally, since most images can be broken down into important features, improvements to our model can generalize to other datasets. Therefore, our model can potentially serve as a superior alternative to traditional means of generating photorealistic images.

## Introduction

Generative modeling is an important part of machine learning as it is used for data synthesis and augmentation. These models automate many manual tasks (e.g. converting satellite images to road maps, and colorizing black and white photos). Implementing a generative model relies on a competition between two separate neural networks that train together. This framework is referred to as generative adversarial networks (GANs) (Goodfellow et al., 2014) and relies on an adversarial game between a generator (model synthesizing data) and a discriminator (model that tries to distinguish real data from synthesized data). Here, the generator tries to fool the discriminator, where the job of the discriminator is to differentiate between real and generated samples. In an adversarial fashion, the discriminator tries to maximize how well it can differentiate between real and fake data. As a result, the discriminator becomes better at classifying real and generated samples, and the generator becomes better at generating samples that can fool the discriminator. With both the generator and discriminator changing during training, it is important to establish an objective metric that does not rely on loss. The Fréchet Inception Distance (FID) measures the distance between the generated distribution and the target distribution (i.e. how realistic the generated samples look) (Heusel et al. 2017). Since the inception of GANs, the image quality of generated results has steadily improved, but the current state-of-the-art GANs are computationally expensive and involve unstable training processes. A common problem in training photorealistic generators, that is generators producing generally higher-quality images than traditional GANs, is that the generated distribution and the model distribution are not similar in higher-dimensional spaces (Zhang et al., 2016). StackGANs is an architecture utilized by Denton et al. (2015) that resolves some of these issues by dividing the training into many simple stages. In their model, each stage of the stacked generators serves as a conditional step in a Laplacian pyramid network generating images of successively higher resolutions. By producing the final resulting image in multiple stages, each stage's complexity is significantly reduced. Starting off at lower resolutions, Denton's stacked architecture can preserve the quality of initially generated images as it progresses through higher resolution stages, resulting in image quality competitive with other state-of-the-art GANs at the final stage. Moreover, each stage can be trained in parallel, using real images at the proper resolutions as conditional inputs. This allows for easier fine tuning of the model, as the problems associated with the overall architecture can be dissected at each individual stage.

Building on the success of this image StackGAN model, we propose an alternative StackGAN solution that focuses on successively increasing feature complexity rather than resolution. Dividing our model into multiple stages, our finalized output will be grayscale facial images.

## Methods

Our model successively generates more complex features as the intermediate image progresses through each stage. Compared to a traditional GAN, the complexity of the target distribution for each individual GAN in our architecture is significantly reduced, allowing for higher quality end results.

In our attempt to optimize the potential of GANs on grayscale images, we decided to train on facial images from the CelebA dataset. Constrained by our compute time on the Texas Advanced Computing Center maverick 2 supercomputer, we reasoned that the ~200,000 facial images in CelebA provides just enough variety to produce meaningful results. Training on these images, we develop a 2 stage StackGAN framework as follows : stage 1 generates edges from random noise and stage 2 generates grayscale from edges (Figure 1). In the first stage, a DCGAN architecture produces edges. In the second stage, the edges are fed into a conditional ResNet-6 architecture, which generates the final grayscale images. However, CelebA isn't prepackaged with grayscale and edges; it only provides RGB images. Thus, we first convert all ~200,000 RGB images into new training sets of both edges and grayscale.

To create these training images, we used the built-in functionality of the Pillow image library (Alex Clark et al., 2021) to convert the original RGB CelebA images to grayscale, which were then converted to edges using the Sobel filter in the Pillow library. (Figure 2). Next, the images are scaled to a dimension of 128x128, which is the input size for our StackGAN model.

Finally, with the images fully preprocessed, we begin the construction, training, and tuning of the model.

At the first stage of our StackGAN, we rely on deep convolutional architecture (DCGAN architecture), which has had great success in producing high quality images at higher resolutions (Radford and Metz, 2016). For the first stage of our StackGAN (noise to edges), we built upon the pre-existing architecture from Kadel (2019) in PyTorch (Paszke et al., 2021) for CelebA edges. From testing, we find that decreasing the number of feature maps of the discriminator by a factor of 4 improves the underlying DCGAN architecture by preventing the discriminator from outpacing the generator. However, even with this alteration, there is still substantial noise in the background of our generated edges.

From the first stage, we use the generated edges as conditional inputs for our second stage GAN (edges to grayscale). For this conditional stage, we utilized the ResNet-6 architecture and trained on a batch size of one to better emulate other state-of-the-art architectures. This maximizes the number of gradient updates and more precisely captures the mapping of each sample. Although this results in only one epoch of training due to limited compute time, we prioritize a smooth convergence in our L1 loss. This is because the conditional generator's priority is to learn the mapping between the input and output images rather than the entire realism of the resulting image.

Lastly, we then used the FID to compare the generated grayscale images of our StackGAN model against established state-of-the-art GAN models trained on CelebA.

## Results

In the first stage of our StackGAN model, we trained on both 50,000 (~25% of CelebA) and 202,599 (100% of CelebA) edges. As depicted in Figure 3, the edges trained on the entirety of the CelebA dataset (rightmost results) have more defined facial features and less background noise. Moreover, the full training results have an FID score of 73 while the partial training results have a score of 181. This strongly indicates that including the entirety of the CelebA dataset in our training is essential to effective image generation due to the greater variety it provides.

Meanwhile, Figure 4 shows the grayscale images generated from both the resulting stage one (left) and real (right) edges as conditional inputs (our model was trained on 202,599 edges for both outputs). The grayscale images generated from real conditional edges have much more defined facial features and substantially less noise throughout the entire image. The FID score for grayscale images generated using synthetic edge images was 59. A FID score for the grayscale images generated using real edges was not calculated since the difference in quality is very apparent.

These sentiments for both stages are further reinforced by human evaluators, who measure the perceived quality of the generator outputs. Although the sample size of human evaluators was quite small, each one was able to easily discern the generated images from the real images in both stages, due to the significant noise present in most outputs. While image quality and diversity are both considered in the FID scoring metric, human evaluators can potentially discern which of these two potential error sources is present, thus enabling us to identify potential problems and appropriate solutions to improve the model. By physically observing the generated images, human evaluators can determine that they were primarily lacking in quality rather than diversity. This problem is further diagnosed and discussed in the following section.

## Discussion

Comparing the generated images (Figure 4) against a sample of the real images (Figure 5), we note that the stage one GAN is successful in defining various facial features (eyes, nose, mouth, and hair). However, the majority of the generated edge samples still contain evidence of noise. The generation of noisy images in the first stage of the StackGAN is not ideal, but it is not a failing point for the architecture. StackGAN baseline stages yield images with the lowest quality. This is due to the first stage generating images from noise instead of building on top of generated images. Therefore, because there is no conditional input to constrain the generated edges from our stage one GAN, it should be expected to produce the least promising results.

Our results also showed that our full model evaluated immediately after the second stage achieved a better FID score than the same model evaluated immediately after the first stage. We believe this is because the conditional stages can take inputs "far away" from training points in the input space, and move them closer to corresponding training points in the output space. In effect, this leads to subsequent stages removing noise from prior stages. More mathematically, we believe this to be true when the inequality $||\vec{f}(\vec{x} + \vec{h}) - \vec{f}(\vec{x})|| < ||\vec{h}||$ is satisfied by the conditional generator, $\vec{f}$, where $\vec{x}$ is the input of a training sample and $\vec{x} + \vec{h}$ is the novel input

generated by the previous stage. Our observation that images improve in quality further down the stack is important, because it demonstrates an area where a stacked architecture can theoretically outperform a single-generator architecture (i.e. in the case of removing noise).

However, there are still ways that these results can be optimized. Using the full CelebA dataset rather than 25% of it resulted in significant reductions in noise. Meaning that more variety in a dataset can increase the generated image quality. The most promising approach to improving the quality of our edge results is to alter the underlying stage one architecture. A promising alternative to the stage one DCGAN is the Wasserstein GAN (WGAN) with gradient penalty. WGANs have been shown to improve learning stability and avoid many problems present in traditional GANs (Arjovsky et al., 2017). In our case, replacing the DCGAN with a WGAN could reduce the oscillation present in the generator and discriminator losses (Figure 6), which would result in higher quality edges from the stage one GAN.

In the stage two (edges to grayscale) GAN, there is a clear difference in quality between the generated and real samples of images (Figures 4 and 5, respectively). This issue in quality appears to be largely affected by noise. However, all the generated grayscale images preserve some of the well-defined features that were generated but with noise artifacts present. This is further reinforced by the lack of convergence in the generator and discriminator loss functions depicted in Figure 7. This could be because we only trained on one epoch. The lack of training for the stage two GAN is due to the increased model complexity of ResNet-6. With the lack of training considered, stage two performed well in learning how to "shade in" the edges produced by stage one. However, the generated face is surrounded by complete blur since the stage one generator doesn't provide the rest of the model any information on background. When it comes to purely discussing the facial features, the grayscale mapping generated from real edges has less noise and more clearly defined features than those generated from the stage one output edges (Figure 4). Furthermore, when comparing the stage two results from real edges and synthetic edges, there seems to be some evidence of overfitting. The stage two outputs from the real edges were near perfect while the novel edges produced considerably noisier and blurrier grayscale images. The inclusion of dropout layers could create a generalized mapping between edges and grayscale and produce higher quality results.

The stage one GAN has a better FID score than the stage two GAN. One possibility is that the output from the stage one GAN adds to the error of the stage two GAN. As a conditional

GAN, the second stage can both strengthen and weaken the given input when translating. While it currently struggles at producing competitive results, it has the potential to improve from the edges. If the edges from the first stage improves, then the grayscale data from the second stage should improve as well. However, the stage two GAN will not show improvement unless the overfit of the mapping is managed. This is because the faulty edge to grayscale mapping has a more severe effect on our model than the moderate noise prevalent in the generated edges.

Given the apparent issues we faced with the flow of information through the overall model (e.g. background information not being passed into the stage two GAN), we propose two new model architectures that aim to mitigate the aforementioned problems (Figure 8). Proposed architecture #1 gives subsequent stages the freedom to fill in gaps using their own latent vectors. This allows the stages to still be trained independently, and it delegates different image features to different stages. For example, the first stage would mostly generate facial features, the second stage would generate background features, etc. This could theoretically produce more diverse output images as different parts of the image would be generated using different latent vectors. Proposed architecture #2 seeks to preserve as much information as possible from the original training images. This is done by using skip connections that relay all prior stage information (including the latent vector) to the current stage. This approach complicates training by forcing all of the stages to train simultaneously, but it provides all stages with the information needed to recreate a realistic image. This contrasts with proposed architecture #1 in the sense that each individual stage may produce results closer to that stage's target distribution, but the stages together produce a result that does not accurately represent the dataset. For example, in proposed architecture #1, a stage could produce a realistic sample of a person snowboarding, and a subsequent stage could produce a realistic sample of a background that appears to feature a sunny day on the beach. These two results alone could be optimal when training stages independently, but in the end they offer nonsensical results when combined. Therefore, if the loss of information problem is fixed by using separate latent vectors, then the latent vectors themselves must be related in some way.

## Conclusion

Our StackGAN allows for more stable training processes and distribution similarity between real and generated data in higher dimensional spaces by dividing its training into multiple simple stages and paving the way for future work in this area, serving a variety of practical generative applications.

Both stages of our StackGAN are susceptible to noise. Replacing our DCGAN architecture with WGAN in the first stage can allow for smoother training and mitigate mode collapse. In the second stage, facial features were significantly more defined than the image background, indicating the presence of noise in the first stage. Including dropout layers serves as a potentially simple optimization to reduce noise and overfitting in the stage two (edges to grayscale) mapping. Also, to mitigate the loss of information between stages, we propose two potential new architectures that train on supplementary conditional inputs per stage. One architecture uses a new latent vector for each stage, and the other uses the outputs of previous stages as inputs for subsequent stages. With the first architecture, it is recommended that the new latent vectors are correlated in some way to prevent the "snowboarding on the beach" problem discussed earlier. The second architecture seems more theoretically sound, but it may be a lot more expensive to train than proposed architecture #1.

Furthermore, our work highlights the importance of training on a diverse dataset, as indicated by the higher quality results obtained when training on ~200,000 CelebA images over 50,000 images. Ultimately, image quality was heavily dependent on the variation present in the dataset, which enabled the generator to produce images with less noise.

Almost every image, regardless of subject, can be broken down into important features (various pieces of disjoint information contained within the image), implying that training can be divided into successive stages. Therefore, given the results we have achieved for facial image generation, our StackGAN model has the potential to generalize to a variety of other datasets.

# Figure Captions

Figure 1: The proposed StackGAN architecture

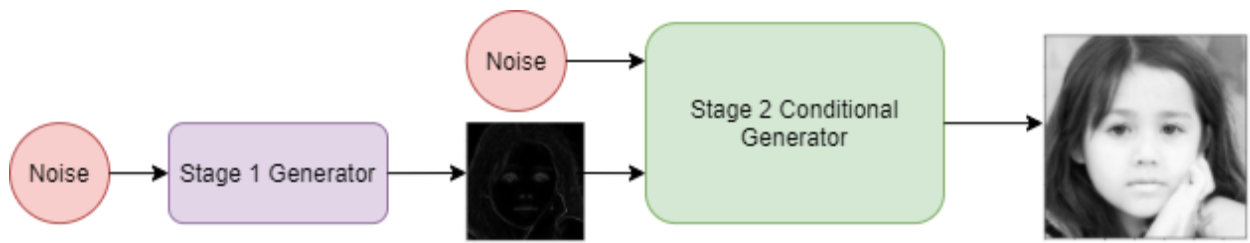

Figure 2: RGB to grayscale to edges transformation on CelebA dataset

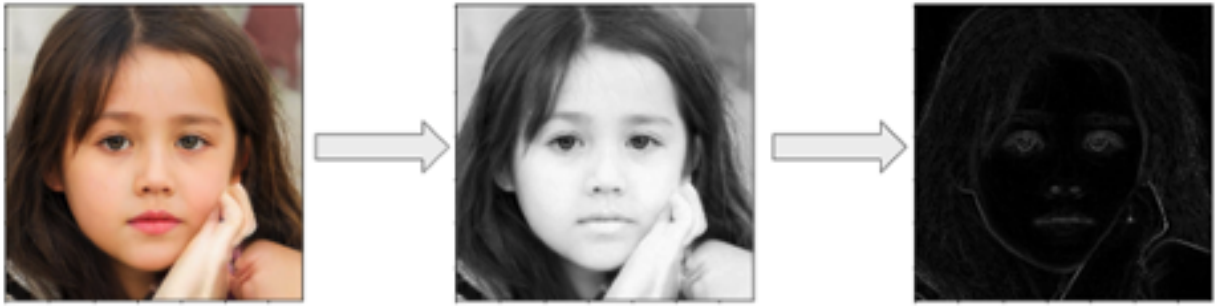

Figure 3: Generated edges using 50k samples (left) vs 200k samples (right)

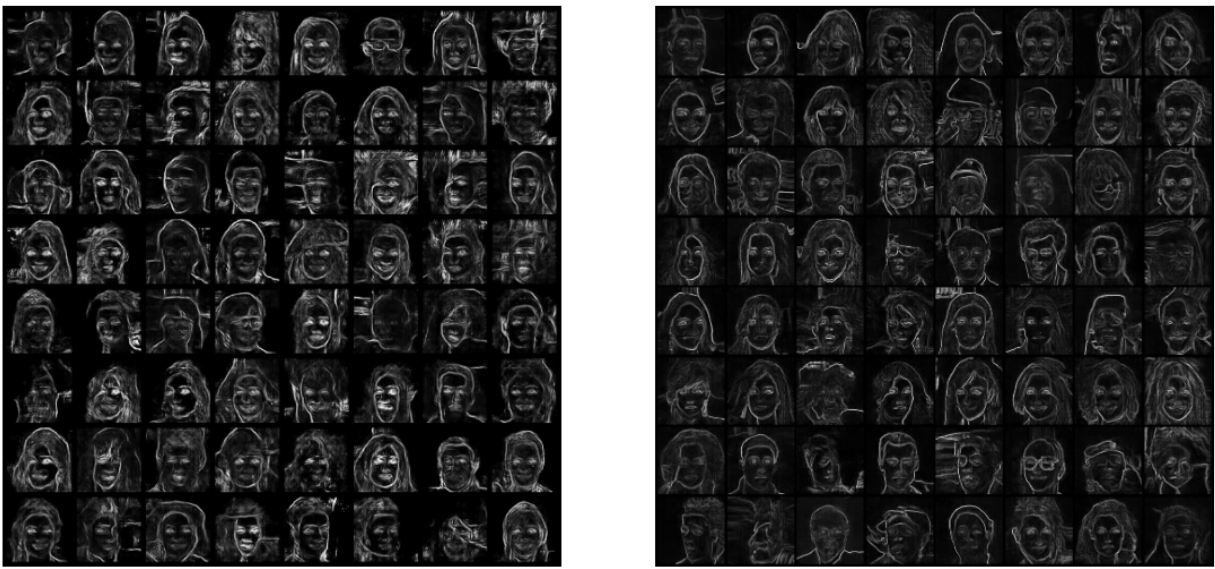

Figure 4: Generated grayscale images using generated edge input (left) vs real edge input (right)

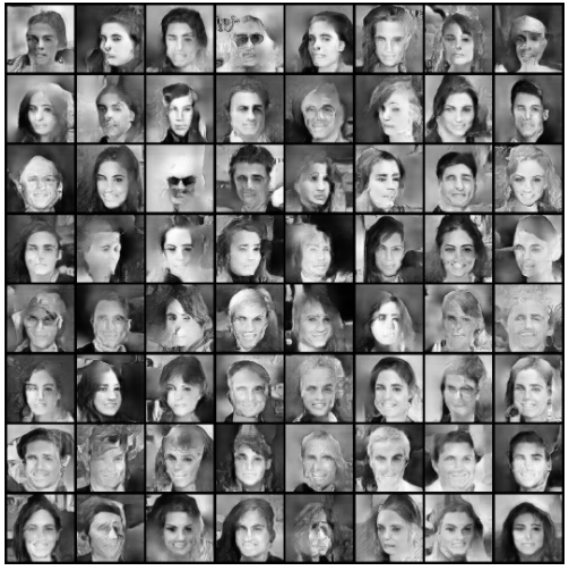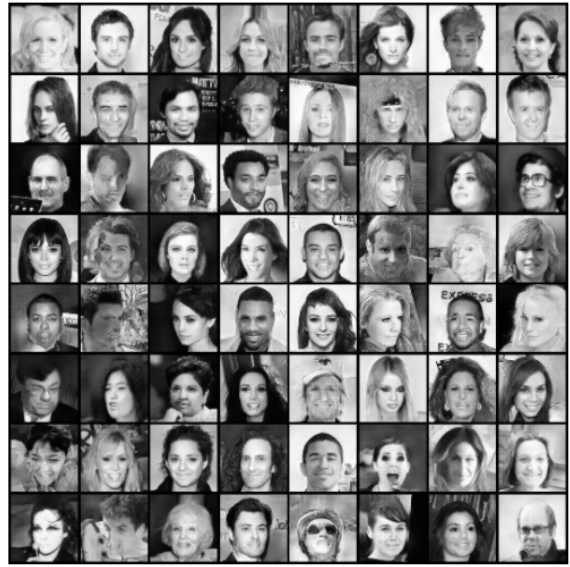

Figure 5: Real grayscale images from the CelebA dataset.

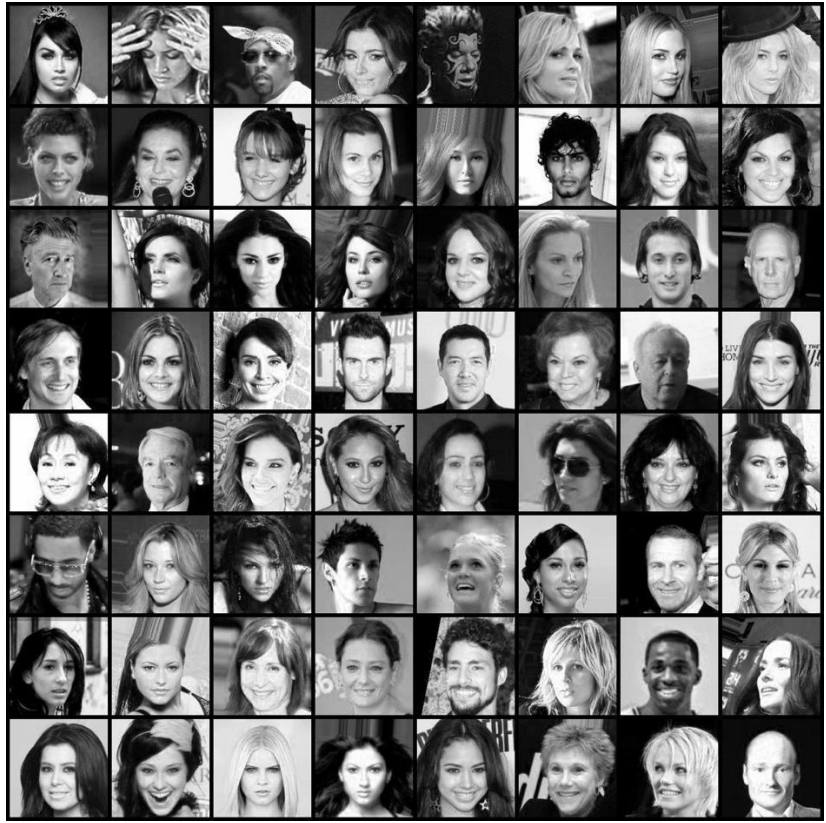

Figure 6: Stage one generator and discriminator adversarial loss during training

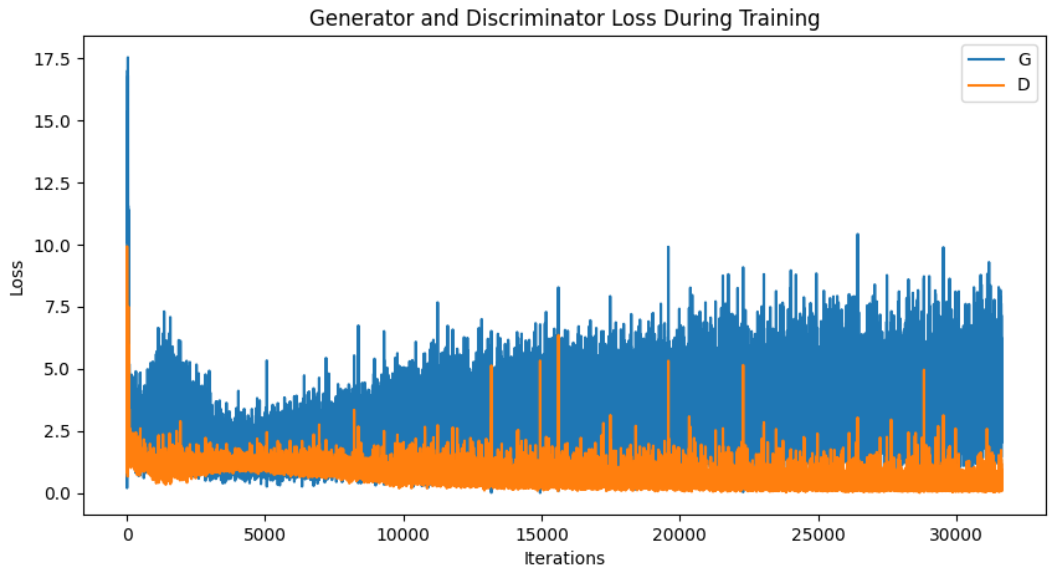

Figure 7: Stage two generator and discriminator loss during training

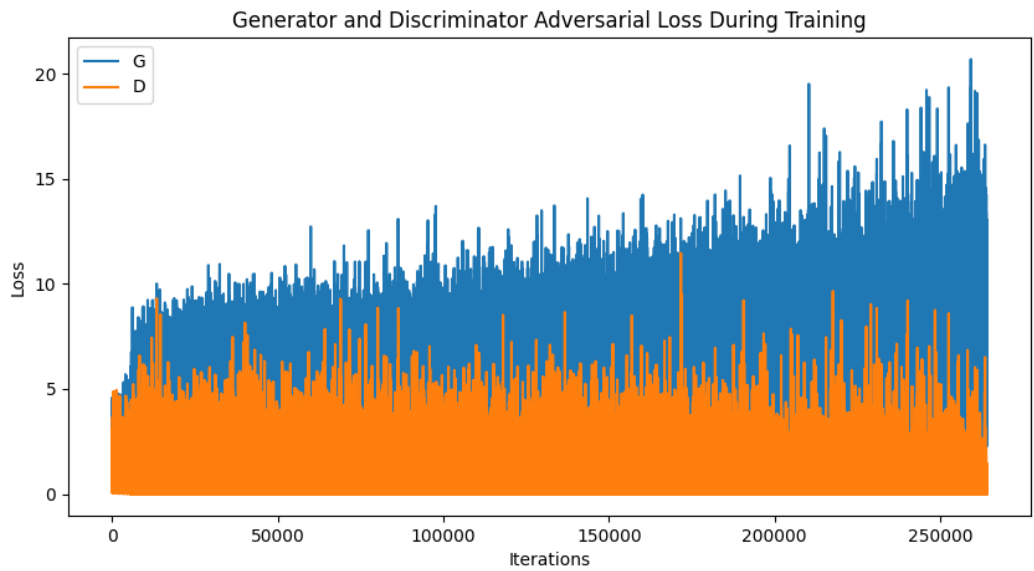

Figure 8: Proposed architectures

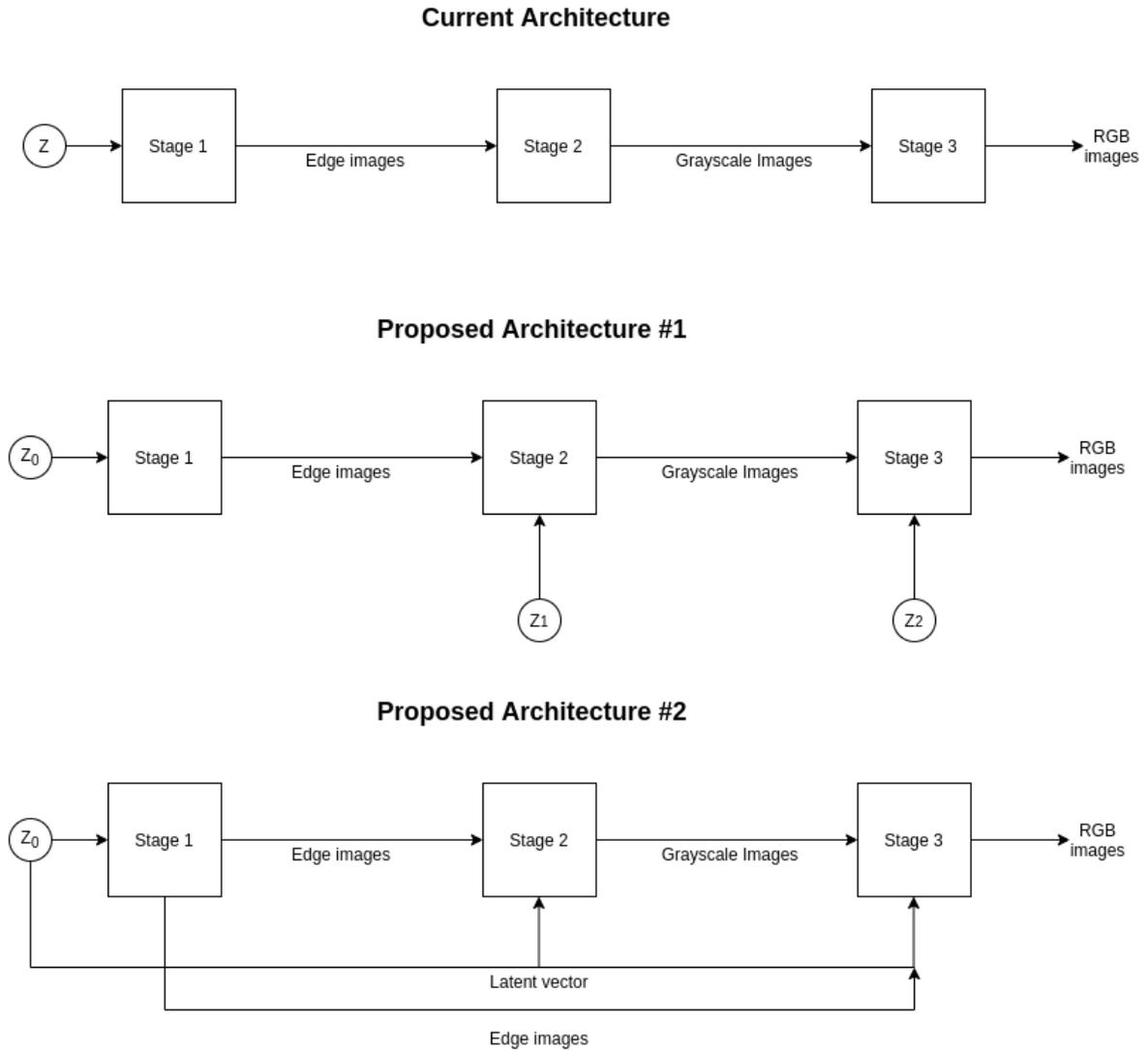

The "Z" prefix denotes a latent vector of arbitrary size.